\documentclass[letterpaper, 10pt, conference]{ieeeconf}

\IEEEoverridecommandlockouts    
\usepackage{times}
\usepackage{url}
\usepackage[utf8]{inputenc}
\usepackage[small]{caption}
\usepackage{graphicx}
\usepackage{amsmath}
\usepackage{booktabs}
\usepackage[linesnumbered,ruled,vlined]{algorithm2e}
\usepackage{nomencl}
\usepackage{wrapfig}
\usepackage{multirow}
\usepackage{tikz}
\usepackage{tabularx}
\usepackage{microtype}
\let\labelindent\relax
\usepackage{enumitem}
\usetikzlibrary{calc, arrows, positioning, shapes, shadows, spy, plotmarks, matrix, fit, backgrounds}

\newcommand{\mvsanet}{{MVSA-Net}}
\newcommand{\yh}[1]{{\color{red}{#1}}}
\begin{document}

\title{\mvsanet: Multi-View State-Action Recognition for\\ Robust and Deployable Trajectory Generation
}

\author{Ehsan Asali, Prashant Doshi, Jin Sun$^{1}$ 
\thanks{$^{1}$Ehsan Asali, Prashant Doshi, and Jin Sun are with the School of Computing, University of Georgia, Athens GA 30606, USA
{\tt\small ehsanasali@uga.edu}
}
\thanks{This work was presented at Deployable AI Workshop at AAAI-2024}
}

\maketitle
\thispagestyle{empty}
\pagestyle{empty}

\begin{abstract}
The learn-from-observation (LfO) paradigm is a human-inspired mode for a robot to learn to perform a task simply by watching it being performed. LfO can facilitate robot integration on factory floors by minimizing disruption and reducing tedious programming. A key component of the LfO pipeline is a transformation of the depth camera frames to the corresponding task state and action pairs, which are then relayed to learning techniques such as imitation or inverse reinforcement learning for understanding the task parameters. While several existing computer vision models analyze videos for activity recognition, SA-Net specifically targets robotic LfO from RGB-D data. 
However, SA-Net and many other models analyze frame data captured from a single viewpoint. Their analysis is therefore highly sensitive to occlusions of the observed task, which are frequent in deployments. An obvious way of reducing occlusions is to simultaneously observe the task from multiple viewpoints and synchronously fuse the multiple streams in the model. Toward this, we present multi-view SA-Net, which generalizes the SA-Net model to allow the perception of multiple viewpoints of the task activity, integrate them, and better recognize the state and action in each frame. Performance evaluations on two distinct domains establish that \mvsanet{} recognizes the state-action pairs under occlusion more accurately compared to single-view \mvsanet{} and other baselines. Our ablation studies further evaluate its performance under different ambient conditions and establish the contribution of the architecture components. As such, \mvsanet{} offers a significantly more robust and deployable state-action trajectory generation compared to previous methods.

\end{abstract}

\section{Introduction}
\label{sec:intro}

Learn-from-observation (LfO) is the human-inspired methodology of observing an expert (usually a human) performing a task and learning a computational representation of the behavior that can then be used by a robotic agent to perform the task as well as the expert~\cite{argall2009survey, Arora19:Survey}. In most methods that enable LfO, the agent learns a mapping between the states of the environment and available actions. This mapping is known as a policy, which the agent later uses to perform the task in the environment. Instead of directly training an end-to-end policy for predicting low-level robotic actions, our approach involves the classification of high-level state-action samples extracted from sensor streams. The main motivation for such a learning paradigm arises from the learner's objective, which may not only be to replicate the expert's actions but also to comprehend the expert's preferences. This comprehension is achieved through the learning of a reward function using state-of-the-art inverse RL methodologies, which, in turn, aids in task acquisition. Consequently, the learner becomes adaptable to novel scenarios, as it has gleaned the expert's preferences instead of adhering to a rigid policy.  

\begin{figure}[t]
\setlength{\belowcaptionskip}{-5pt}
\centering
\begin{tikzpicture}[thick, spy using outlines={rectangle,lens={scale=2.5}, width=1cm, height=1.5cm, connect spies}]
	\node (reg_id1) {\includegraphics[width=1.0\columnwidth]{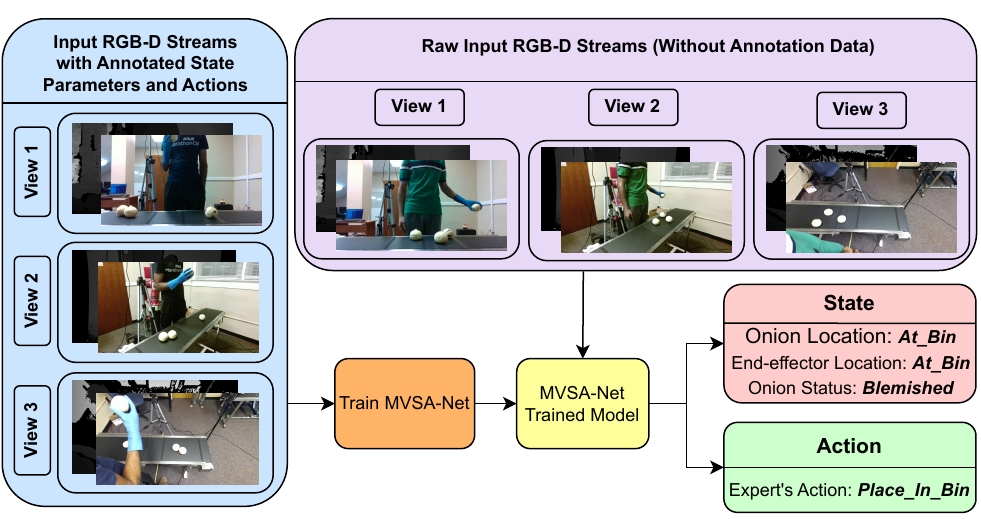}};
\end{tikzpicture}
\caption{An overview of how to use \mvsanet{} for trajectory generation in a custom domain (onion-sorting) having three input RGB-D streams. First, synchronized RGB-D frames are annotated and used to train the state and action classifiers. Later, unseen synchronized RGB-D frames can be fed to the trained \mvsanet{} model to predict the state parameters and the expert's action. In the above example, the expert is about to place a blemished onion in a cardboard box (bin) which is located behind the conveyor belt.}
\label{fig:summary}
\vspace{-4mm}
\end{figure}

In order to bridge the gap between the raw vision (often RGB-D) sensor stream input and the state-action samples as specified by the environment model, several previous computer vision models exist that analyze videos for activity recognition~\cite{amirian2018dissection, islam2022human}, and there are analogous ones for robots that analyze depth camera streams~\cite{cortes2012using, fan2016robust, graeve20143d}. A particular deep neural network model specifically aimed for robotic LfO from RGB-D data is SA-Net~\cite{soans2020sa}. A trained SA-Net model processes raw RGB-D streams from a single depth camera and produces joint state-action trajectories. 
SA-Net's convolutional layers utilize the RGB channels to extract spatial information and identify the state, while time-distributed CNNs and LSTMs~\cite{LSTM} handle action recognition. To recognize actions, the network extracts spatial and temporal information to predict motion direction. Nonetheless, this limitation to a single viewpoint comes with well-known challenges. A perception sensor's field of view (FoV) determines the span of its visibility, limiting the information it gathers. Sensor streams exhibit various levels of hardware noise, characterized by their signal-to-noise ratio. 

Finally, unavoidable occlusions in the environment could block the view of the camera. To illustrate this, consider the use case of LfO of a human sorting produce on a conveyor in a farm-based packing shed. The wall-mounted depth camera sensor may get irregularly obstructed by other line workers, humans wishing to converse with the observed human, or small vehicles moving about on the shed floor~\cite{Bogert22:Hierarchical,Suresh22:Marginal}. LfO paradigms and specifically IRL techniques are highly susceptible to occlusion and noise, while these challenges are common in deployed environments. Our multi-view model is best suited in such cases where the chance of occlusion and/or noise is high or when the domain is sensitive and losing track of states/actions can harm the system significantly. Furthermore, a single sensor is insufficient to capture both the actions and what the human sees as it inspects the produce. 

In this paper, we present \mvsanet{}, a model that enables input from multiple viewpoints and recognizes state-action pairs from multiple synchronized RGB-D data streams. \mvsanet{} is a generalized model that leverages multiple heterogeneous visual sensors to simultaneously identify the expert's states and actions in challenging real-world scenarios. 
For our produce processing use case, multiple depth cameras can allow both the expert's actions and his or her view of the product during inspection to be recorded; multiple viewpoints also help mitigate the detrimental effects of occlusions and sensor noise on LfO. \mvsanet{} utilizes gating networks~\cite{chung2015gating} to establish a mixture-of-experts layer, where several expert models represented by input streams make classification choices and gating controls how to weigh the classifications in a probabilistic input-dependent manner. Thus, the gating dynamically distributes the decision-making to the different views for optimal performance. The model employs time-distributed CNNs and GRU recurrent units, receiving frames from the current time step $t$ and the four previous time steps to create a temporal context that facilitates action recognition.

We conduct a comprehensive performance evaluation of \mvsanet{} in two robotic domains. In the first domain, we illustrate the produce-sorting task, where a human expert sorts onions, retaining unblemished onions on the conveyor while removing blemished ones upon inspection. 
In this scenario, the model processes three RGB-D camera streams concurrently to identify the state, including the onion's location, the end-effector's (sorter’s hand) position, the onion status, and the expert’s actions. 
We employ a custom-trained YOLOv5 instance~\cite{redmon2016you} to classify the onion. 
The second domain involves mobile robot patrolling, as introduced in~\cite{bogert2014multi}, where two TurtleBots patrol a hallway simultaneously but independently. Another TurtleBot observes the patrollers from a vantage point and attempts to penetrate the patrol. To assist the attacker, we install an additional camera at a second vantage point in our experiments. The attacker utilizes \mvsanet{} to recognize patrollers’ movements.  
In both domains, \mvsanet{} demonstrates a significant improvement in state-action prediction accuracy when compared to single-view and multi-view baselines. Furthermore, our comprehensive ablation studies validate \mvsanet{}'s effectiveness and robustness in challenging scenarios, such as malfunctioning sensors and varying lighting conditions. These ablation studies also empirically establish the necessity of specific modules in the model's architecture. \mvsanet{} is, to our knowledge, the first model for multiple viewpoint state-action pair recognition, facilitating robot LfO. 

\section{Related Work}
\label{sec:related}

Recent advances enabled by deep neural networks have significantly enhanced the performance of image and video classification tasks~\cite{DBLP:journals/corr/HeZRS15,yue2015beyond}. A substantial body of work has predominantly concentrated on single-view human activity and action recognition~\cite{Rezazadegan17:Action,DBLP:journals/corr/WangLGZTO17,soans2020sa}. Although these techniques excel under normal conditions, they are susceptible to environmental occlusions and sensor noise in the data. 

Multi-modal and multi-view sensor input in the context of classification problems has been shown to overcome the limitations of single-view or single-modality systems. For instance, Asali et al.~\cite{asali2021deepmsrf} utilize visual and audio streams within a deep multi-modal architecture to conduct speaker recognition in video clips. 
Shin et al.~\cite{shin2019context} introduced a context-aware collaborative pipeline for multi-view object recognition. The pipeline extracts features from each view at each time step, applies view pooling to the aggregated result, and conducts a global prediction for each frame's object class. 
Lu et al.~\cite{lu2021custom} present a multi-camera pipeline that detects objects in each camera stream using shared pseudo-labels and performs a consistency learning step. Kasaei et al.~\cite{kasaei2021simultaneous} developed a deep learning architecture equipped with augmented memories to concurrently address open-ended object recognition and grasping. This method accepts multiple views of an object as input and jointly estimates pixel-wise grasp configurations and scale-and-rotation-invariant feature representations as outputs. 
Strbac et al.~\cite{strbac2020yolo} put forward a YOLO detector model and the principles of stereoscopy to substitute the Light Detection And Ranging (LiDAR) sensors with cameras for distance estimation. In~\cite{martin2018enhancing}, Nieto et al. introduce a multi-camera people detection framework that employs a camera transference unit to transfer detection results from auxiliary cameras to the primary camera. 
This method can transfer detections to a different point of view, automatically combine multiple cameras, and automatically select the working threshold for each of them. 

Some works investigate the utilization of multi-view~\cite{vyas2020multi,liu2021mlrmv,wang2018dividing} or multi-modality~\cite{ma2019attnsense} data in action recognition tasks. For instance, Vyas et al.~\cite{vyas2020multi} suggest an unsupervised representation learning framework that encodes scene dynamics in videos captured from multiple viewpoints and predicts actions from unseen views using RGB data. Nevertheless, this method is susceptible to performance degradation in the presence of noise in one or multiple views.
Additionally, it employs multiple views during training but only a single view during testing.
In contrast, \mvsanet{} learns the states and actions of a person from each viewpoint and subsequently predicts the state-action pair from multiple views. 
Wang et al.~\cite{wang2018dividing} propose a multi-branch network called DA-Net for multi-view action recognition. DA-Net exchanges messages among view-specific features from different branches and employs a view-prediction-guided fusion method to produce combined action classification scores. 
While these methods focus on multi-view or multi-modality object-detection or action recognition,  \mvsanet{} simultaneously recognizes state-action pairs for LfO tasks.

To explicitly model temporal dependencies in video frames for action recognition, a few methods have used long short-term memories (LSTMs)~\cite{ullah2017action, wang2016beyond}. LSTMs, as part of the recurrent neural networks (RNNs) family, extract dynamic features from videos. A more efficient and simpler RNN architecture is the gated recurrent unit (GRU). Both GRUs and LSTMs can learn dynamic video features and mitigate the vanishing gradient problem, a common issue in deep neural network training. In \mvsanet{} we choose GRUs over LSTMs due to their simpler structure and faster training characteristics. Additionally, GRUs perform well with video frames where actions occur over short durations, aligning with our task setup. 

\begin{figure*}[t]
\setlength{\belowcaptionskip}{0pt} 
\setlength{\belowcaptionskip}{-15pt}
\centerline{\includegraphics[width=1.0\textwidth]{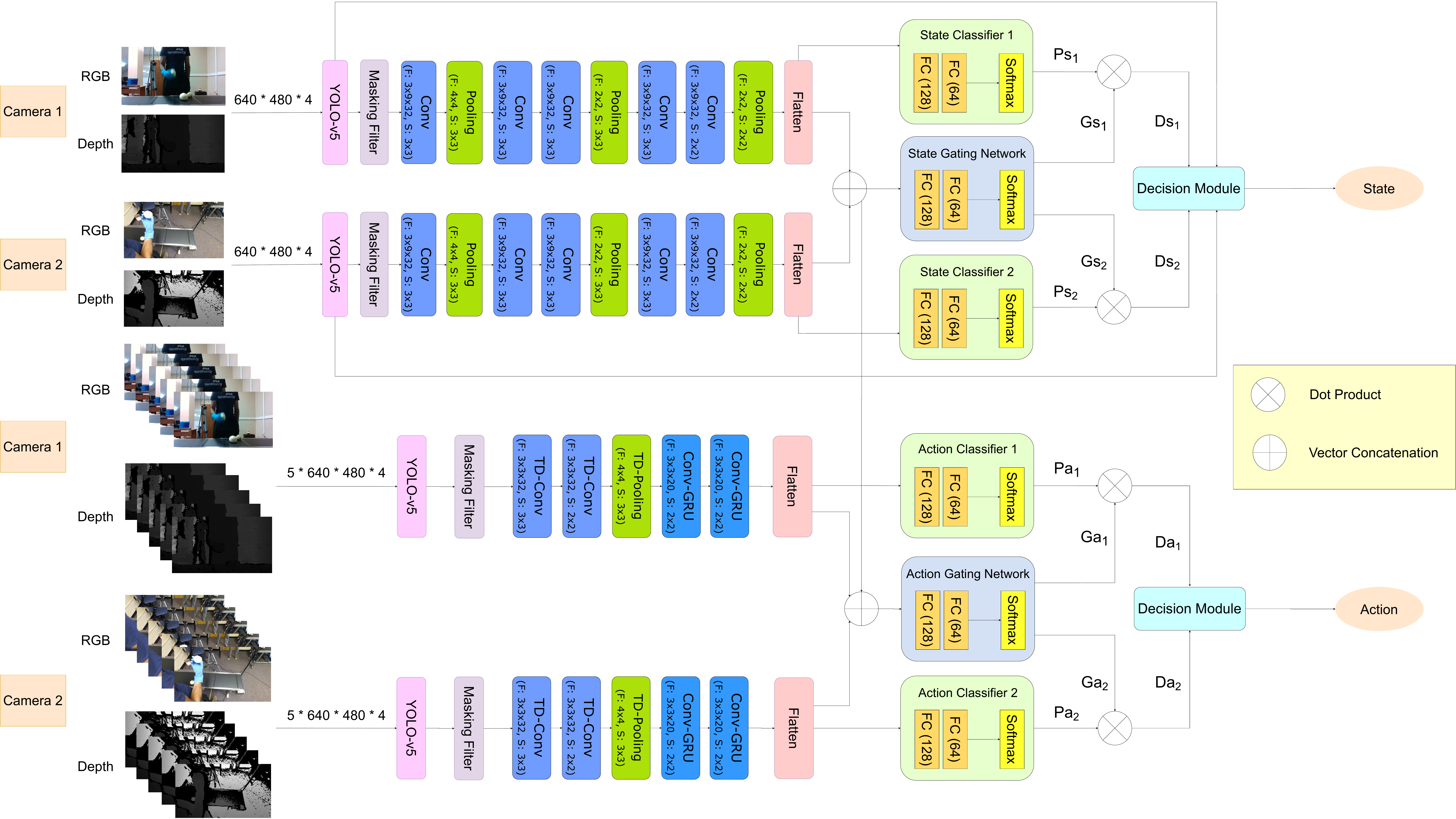}}
\caption{An overview of the \mvsanet{} architecture. The network jointly predicts the state and action of an expert using multiple heterogeneous synchronized RGB-D frames. The so-called RGB-D frames are processed through a deep convolutional and recurrent neural network, which predicts both the state labels and the expert's actions. YOLO-v5 contributes to the network's masking paradigm and aids the decision module in achieving more accurate state predictions. The masking paradigm functions such that when it detects more than one expert in a frame, it automatically converts the input frame into multiple frames, each containing only a single expert.} 
\label{fig:mmsanet}
\end{figure*}

\section{Multi-View SA-Net}
\label{sec:method}

In this section, we provide a detailed description of \mvsanet{}'s architecture. It effectively utilizes both convolutional and recurrent neural network modules and allows for a multitude of input streams. Its adaptable design facilitates easy customization for a wide range of application domains.

\subsection{Method Overview}

We are interested in the problem of estimating the states and actions of experts performing a task from multiple heterogeneous sensors, including RGB-D cameras. The task can be performed by one or multiple humans and/or robots, either real or simulated, who are capable of performing it optimally. We propose to use convolutional and recurrent deep neural networks to analyze the sensory streams from the heterogeneous sensors and extract spatial and temporal information. This information is then employed for recognizing the states and actions of the experts performing the task.
To be specific, given $n$ number of consecutive video frames of an expert performing a task (time steps $t$, $t-1$, ..., $t-(n-1)$), a learner utilizes \mvsanet{} to jointly predict the state and action at the current time step $t$. In our experiments, we set $n$ to 5 as it helps to recover sufficient information from previous time steps to recognize the performed action with relatively high accuracy.  
Fig.~\ref{fig:mmsanet} illustrates \mvsanet{} configured with two cameras/views. \mvsanet{} is adaptable and capable of accommodating any number of views (limited to the hardware capacity), depending on the specific application domain. 
We discretize the state and action spaces to formulate the learning task as a classification problem. 

In a general context, each view can be either a uni-modal (i.e. either RGB or depth) or a multi-modal (RGB-D) video stream. In \textbf{state recognition}, the RGB and depth frames from the most recent time step undergo processing through a series of convolutional and pooling layers. 
The final pooling layer's output is flattened into a 1-D vector and subsequently fed into a neural network known as the state classifier, comprising two hidden layers and one output layer. 
The state classifier network is a multi-head neural network that calculates the probability distribution for each state parameter based on the input frame (i.e., the number of head/output nodes equals the number of state parameters). 
The gating network combines the flattened layers from all views into a single input and generates a probability distribution across the viewpoints. The gating network's architecture mirrors that of the state classifier, with the key difference being that it outputs the probability of each view making a correct prediction about the states (i.e., the number of head/output nodes matches the number of viewpoints). 

The \textbf{action recognition} branch follows a similar process, except that it retrieves \yh{$n$} consecutive video frames from each video stream and employs Time Distributed (TD) convolutions and GRU layers to extract crucial temporal features. 
The gating network utilized for action recognition not only leverages the data from the GRU units but also integrates the aggregated information from the flattened layers of state recognition. The intuition is that the current action and state are strongly correlated. In both the state and action recognition tasks, a decision module is employed to gather the output results from each viewpoint and the gating network, ultimately generating the final output. 

In the following subsections, we describe the state and action recognition mechanisms in full detail, respectively.

\subsection{State Recognition}
The network architecture for generating the state parameters is simple yet efficient. It starts by fetching the latest RGB-D frame from each view, concatenating them into a single multi-dimensional tensor, and subsequently inputting this combined data into a sequence of five convolutional layers and three pooling layers.  
Each convolutional layer uses 32 filters with a kernel size of $3 \times 9$ and a stride of $3 \times 3$, with the exception of the final convolutional layer, which uses a stride of $2 \times 2$. The pooling layers are located after the first, third, and fifth convolutional layers, respectively, with pooling sizes of $4 \times 4$, $2 \times 2$, and $2 \times 2$ and strides of $3 \times 3$, $3 \times 3$, and $2 \times 2$, respectively. To ensure stable training, batch normalization is implemented after the fifth convolutional layer, following the methodology outlined in~\cite{ioffe2015batch}. Since the state features are not a temporal construct and they represent spatial locations of objects as well as visual object specifications, stacking multiple observations would be redundant for state recognition. 

\begin{figure}[!ht]
\setlength{\belowcaptionskip}{-5pt}
\centering
\begin{tikzpicture}[thick, spy using outlines={rectangle,lens={scale=2.5}, width=1cm, height=1.5cm, connect spies}]
	\node (reg_id1) {\includegraphics[width=0.9\columnwidth]{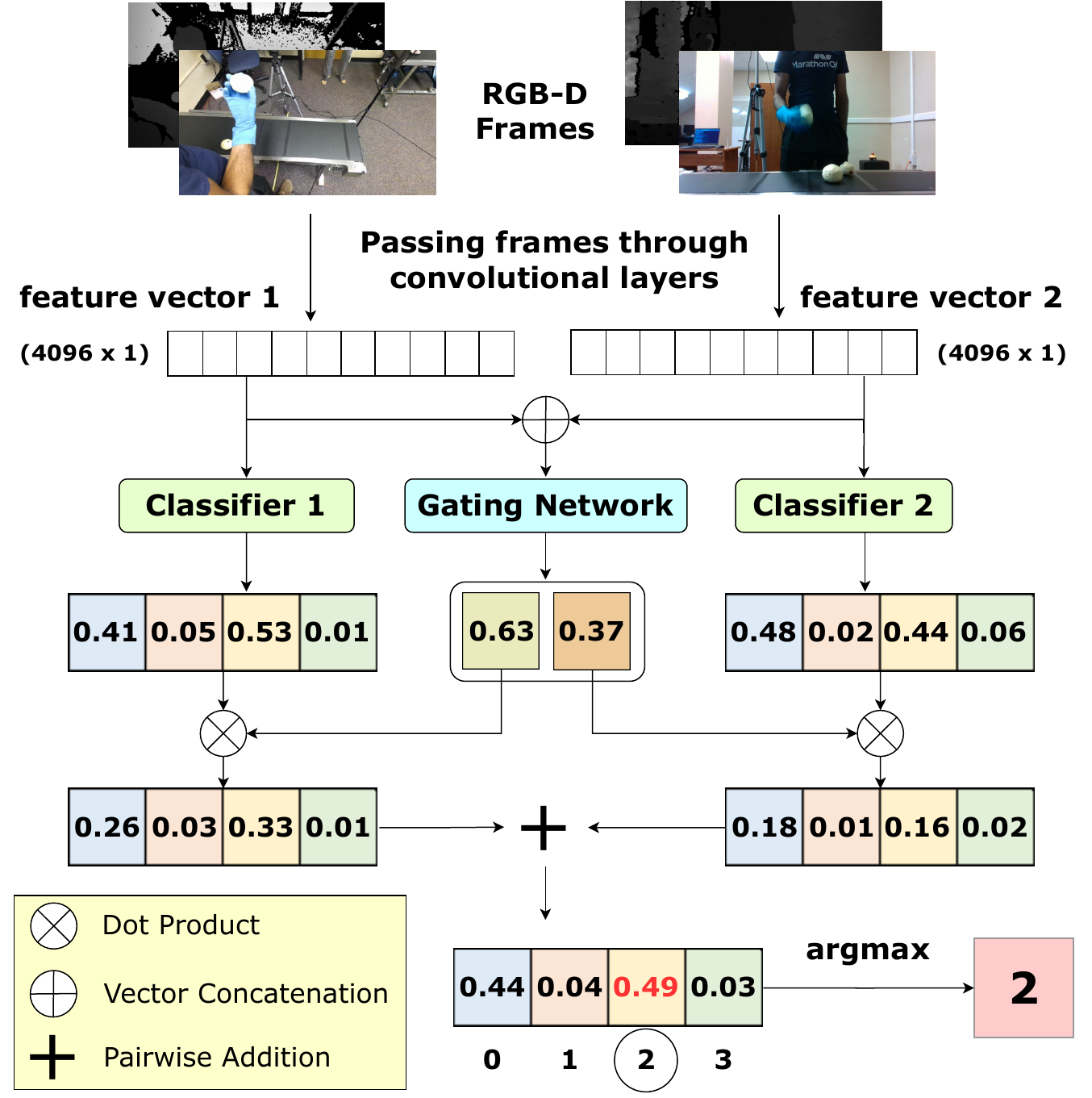}};

\end{tikzpicture}
\caption{A comprehensive example of how the gating network consolidates the results of two distinct classifiers into a solitary output value within the onion sorting domain for recognizing the onion location. The onion location comprises four possibilities, namely $at\_home$, $on\_conveyor$, $in\_front$, and $at\_bin$. Each of these states corresponds to an index ranging from 0 to 3, respectively. As depicted in the diagram, the gating network recognizes that classifier 1 provides more accurate predictions in these scenarios and assigns it a higher weight.}

\label{fig:gating}
\end{figure}

After applying convolution, pooling, and batch normalization operations, the resulting tensor is flattened into a one-dimensional feature vector. Each feature vector ($f_n$) serves as input for the state classifier specific to its corresponding viewpoint ($C_n$). Meanwhile, the gating network takes in the aggregated feature vectors from all views ($f$) and generates a probability distribution across the viewpoint branches ($G_n$). Then, the decision module takes the weighted outputs of each classifier and performs a pairwise addition to unify the vectors. Finally, the output label ($L$) index is obtained by applying the argmax function to the unified vector, as demonstrated in Eq.~\ref{gating_equation}. 

\begin{equation}
\begin{aligned}
L &= \operatorname*{argmax}\left(\sum_n G_n(f)\cdot C_n(f_n)\right)
\end{aligned}
\label{gating_equation}
\end{equation}

The design of the gating network involves assigning a lower weight to the feature vector of a noisy or less informative view and assigning a higher weight to the feature vector of a useful view for predicting the label. The dynamic weighting of diverse views empowers the gating network to optimize each view's contribution and improve the effectiveness of the recognition task. Fig.~\ref{fig:gating} provides a detailed depiction of how the gating network operates within a real-world LfO domain.

\subsection{Action Recognition}
\label{sec:action}
In addition to recognizing the expert's state, \mvsanet{} also estimates the corresponding action at the current time step.
As mentioned earlier, we have designed a network module comprising two time-distributed convolutional (TD-Conv) layers and two GRU layers to accomplish this. Additionally, a time-distributed max pooling layer is also incorporated for sub-sampling. These layers collect image features necessary for action recognition from multiple time steps ($t-4$, $t-3$, $t-2$, $t-1$). Each TD-Conv layer consists of 32 filters with a size of $3\times 3$ and stride of $2\times 2$, while the pooling layer employs a filter of size $4\times 4$ and stride of $3\times 3$. To further compose higher-level features and capture temporal changes in the image sequence, two convolutional GRU layers are employed. Each of these two GRU layers is equipped with 32 kernels sized at $3\times 3$ and stride of $2\times 2$. 
Similar to state recognition, we apply batch normalization for action recognition. As spatial information is crucial for correct action determination, we grab extracted features from the last time step of the state recognition stream. The action recognition branch, like the state recognition one, incorporates a gating network and decision module to optimize performance.

In \mvsanet{}, Negative Log Likelihood~\cite{nguyen2020negative} and Adam~\cite{kingma2015adam} serve as loss and optimization functions, respectively. Section~\ref{sec:experiments} evaluates \mvsanet{'s} performance and compares it to baselines across various conditions.

\section{Experiments}
\label{sec:experiments}

We demonstrate the generalizability of \mvsanet{}'s architecture for LfO by evaluating in two tasks across different domains. One task involves the robotic automation of processing line tasks, while the other focuses on mobile robot patrolling. 
We will provide brief descriptions of these domains, followed by our experimentation procedures and results.

\subsection{Domain Specifications}
\label{sec:domain}

\begin{figure}[!ht]
\centering
\begin{tikzpicture}[thick, spy using outlines={rectangle,lens={scale=2.5}, width=1cm, height=1.5cm, connect spies}]
	\node (reg_id1) {\includegraphics[width=0.98\columnwidth]{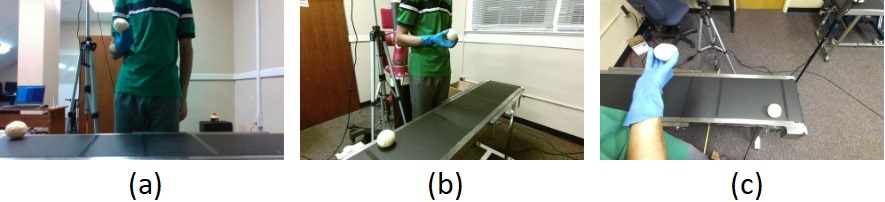}};
\end{tikzpicture}
\caption{Multiple viewpoints of onion sorting: (a) front, (b) side, and (c) top-down views.}
\label{fig:view}
\vspace{-4mm}
\end{figure}

\noindent\textbf{Robot integration into produce processing.}~ In this domain, a robot is tasked with observing and learning how a human sorts an arbitrary number of onions: separating blemished onions from unblemished ones. The human expert picks up onions from the conveyance and meticulously inspects them. If an onion is recognized as blemished, it gets discarded into a bin positioned next to the conveyor belt. Conversely, if it is deemed unblemished, the expert returns it to the conveyance and proceeds to inspect the next onion until all onions are processed.
In our setting, the robot utilizes three heterogeneous depth cameras with different viewpoints. 
Specifically, we utilize an Intel RealSense D435i camera positioned in front of the expert, a Microsoft Kinect V2 camera on the side, and a Microsoft Azure camera above the expert's shoulder.
The latter camera offers a near-point-of-view perspective of the expert's actions.
Fig.~\ref{fig:view} displays the views captured by these three cameras in this task. 

The learner utilizes \mvsanet{} to recognize both the states of the environment and the expert's actions throughout the task. After obtaining trajectories from \mvsanet{}, a robotic manipulator equipped with a gripper can learn the task parameters through imitation learning or inverse RL and subsequently execute the sorting. For the onion-sorting experiment, we collected a dataset consisting of 2,873 synchronized RGB-D frames.

\begin{figure}[!ht]
\centering
\begin{tikzpicture}[thick, spy using outlines={rectangle,lens={scale=2.5}, width=1cm, height=1.5cm, connect spies}]
	\node (reg_id1) {\includegraphics[width=0.95\columnwidth]{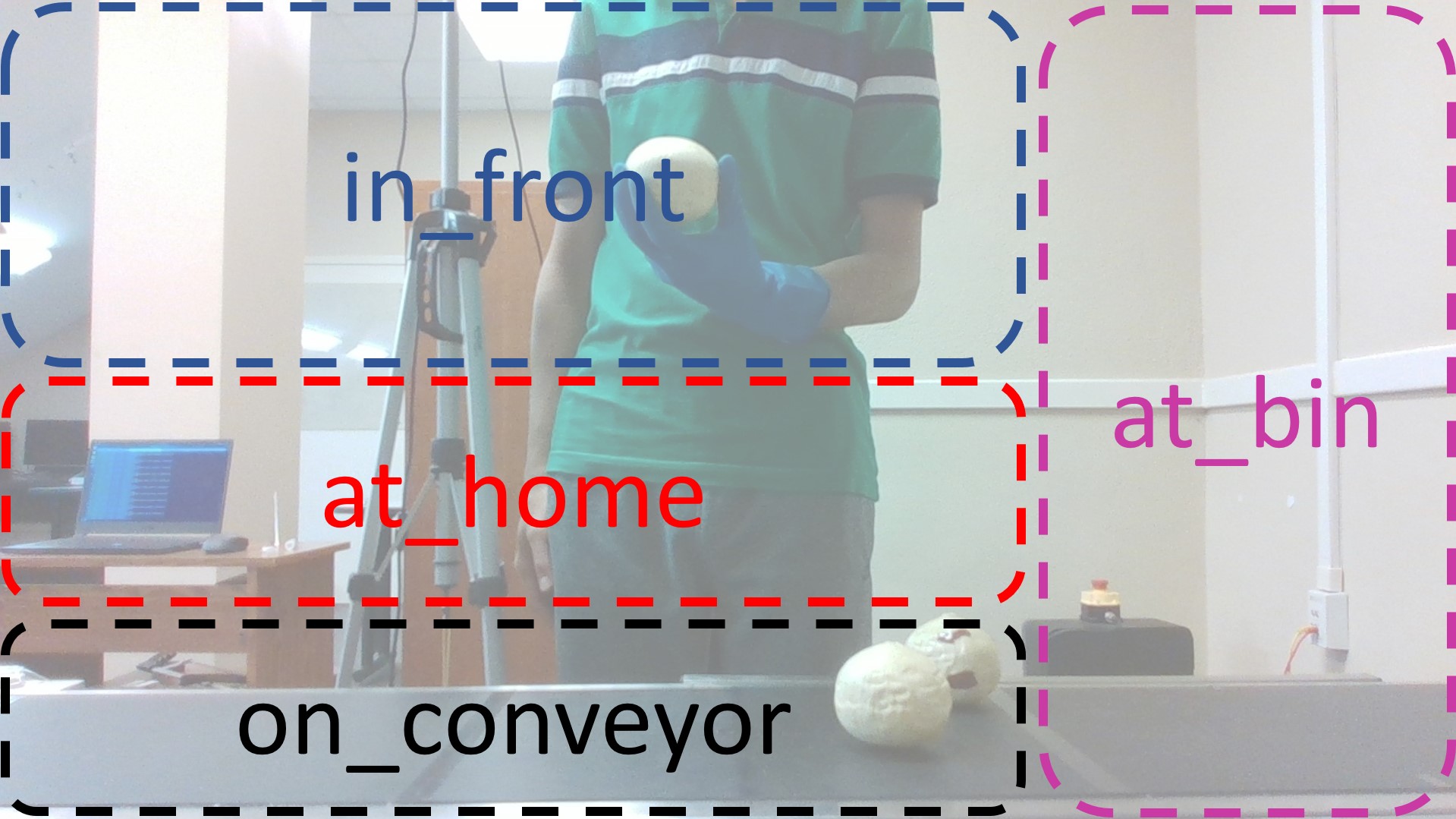}};
\end{tikzpicture}
\caption{Pre-defined spatial regions for the onion and end-effector locations from the front-view camera. In this example, onion and end-effector locations are both in the $in\_front$ region. Furthermore, the onion status is $unblemished$ and the action is $inspect$.}
\label{fig:state_regions}
\end{figure}

The state can be adequately represented by three variables: the spatial location of the onion, the spatial location of the end-effector (hand), and whether the onion is blemished or not.  
Since the camera positions remain stationary, temporal information may not be necessary for state recognition, and the current frame alone suffices. 
\mvsanet{} employs convolutional and pooling layers for recognizing the onion and end-effector locations. Simultaneously, the onion's status is determined using an auxiliary object detector (YOLOv5). This information is then fed into \mvsanet{}'s decision module to create a comprehensive state representation.
We segment the RGB frame into four exclusive spatial regions ($on\_conveyor$, $at\_home$, $in\_front$, and $at\_bin$), as shown in Fig.~\ref{fig:state_regions}, to turn the end-effector and onion location recognition into a classification problem with a finite number of classes. Also, the status of each onion can be either $blemished$, $unblemished$, or $unknown$. Thus, 48 state classes (combinations of end-effector locations, onion locations, and onion status) are possible here. The action classes include $claim$, $inspect$, $place\_in\_bin$, and $place\_on\_conveyor$. Note that each training frame is assigned with a state and action class.
For instance, the ground truth state label for the frame shown in Fig.~\ref{fig:state_regions} is ($in\_front$, $in\_front$, $unblemished$) representing end-effector location, onion location, and onion status, respectively; and the action class is $inspect$.

Multiple onions usually appear in a single frame. To focus YOLO's blemish recognition on the onion currently under inspection, we establish a set of conditions to achieve the desired outcome. 
Prior to assessing these conditions, all onion detections are arranged in descending order based on their classification confidence. Subsequently, each candidate bounding box and its predicted label are taken into consideration. The conditions are defined within a decision module in the final step of \mvsanet{}: 
\begin{itemize}[leftmargin=*, topsep=0in, itemsep=0in]
  \item If no onion is detected with classification confidence of 0.5 or higher, the status is marked as $unknown$. 
\item If the Euclidean distance between the onion and the end-effector is less than 40 pixels, the onion's label is returned. 
  \item Otherwise, the label of the closest onion to the end of the conveyor belt is chosen, and others are disregarded. 
\end{itemize}

{\em Lastly, we need to consolidate predictions from various viewpoints into a single prediction for the status of each onion.} 
It's important to note that a blemished onion may only exhibit the blemish from one viewpoint. Consequently, we establish conditions in the decision module to determine the final label for the targeted onion: If all detections are $unknown$, the onion's status remains $unknown$. However, if at least one viewpoint detects a blemish, then the status is classified as $blemished$; otherwise, it is categorized as $unblemished$.
If all detections are $unknown$, then the onion's status is $unknown$. If at least one viewpoint detects a blemish, then the status is $blemished$, otherwise, the status is $unblemished$.

\begin{figure}[!ht]
\vspace{-2mm}
\centering
\begin{tikzpicture}[thick, spy using outlines={rectangle,lens={scale=2.5}, width=1cm, height=1.5cm, connect spies}]
	\node (reg_id1) {\includegraphics[width=0.95\columnwidth]{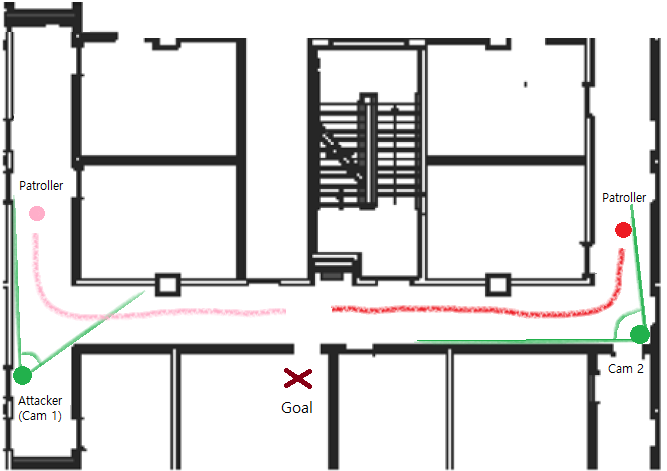}};
\end{tikzpicture}
\caption{The floor map in our patroller-attacker domain. The attacker is equipped with two Kinect V2 cameras (Cam 1 and Cam 2), each with a limited field of view shown in green. The initial position of the patrolling robots and the goal of the attacker is also depicted.}
\label{fig:boyd}
\vspace{-3mm}
\end{figure}

\noindent\textbf{Penetrating robotic patrolling.}~In this domain, an attacker (a TurtleBot2 with an RGB-D camera) observes the movement of two independent patrolling robots (also TurtleBots) from two vantage points. The attacker's objective is to reach a predetermined location without being spotted by the patrolling robots, which have a limited field of view~\cite{bogert2014multi}. 
In this scenario, the attacker employs \mvsanet{} to recognize the states and actions of the patrollers. The state parameters are the poses ($X$,$Y$, and the orientation $\theta$) of the patrolling TurtleBots, with coordinates derived from a 2D map of a hallway (refer to Fig.~\ref{fig:boyd}) where the patrollers travel and the orientation is one of four directions, $north$, $south$, $east$, and $west$. The action classes here include $move\_forward$, $turn\_right$, $turn\_left$, and $stop$. If YOLOv5 fails to detect the TurtleBot2, the network classifies this as $unknown$ for all state and action parameters. 
We employ the same configuration as in prior applications of this domain~\cite{arora2019online,soans2020sa}, with the exception that in our experiment, the attacker utilizes an auxiliary camera to gain an extra perspective of the environment. 

\begin{figure}[t]
\centering
\begin{tikzpicture}[thick, spy using outlines={rectangle,lens={scale=2.5}, width=1cm, height=1.5cm, connect spies}]
	\node (reg_id1) {\centerline{\includegraphics[width=0.95\columnwidth]{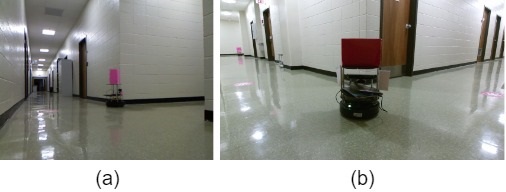}}};
	
\end{tikzpicture}
\caption{Two different viewpoints of the attacker TurtleBot2.\\ ($a$) primary view (Cam 1); ($b$) supplementary view (Cam 2).}
\label{fig:attacker_view}
\vspace{-2mm}
\end{figure}

Note that in comparison to the previous domain, \mvsanet{} recognizes the state-action pairs of multiple experts instead of a single expert, for which the masking filter is activated. 
Additionally, Cam 2 is positioned on the opposite side of the hallway to provide the attacker with a view of both the long and short hallways.
Fig.~\ref{fig:attacker_view} depicts the physical setup and the field of view of the two cameras used by the attacker. The dataset gathered in this domain comprises 5,110 synchronized RGB-D frames captured by each of the two attacker cameras, with a consistent frame rate of 10 FPS. 

\subsection{Baselines}
\label{sec:baseline}

We assess \mvsanet{}'s performance in comparison to three baseline methods: 
\begin{enumerate}[leftmargin=*, topsep=0in, itemsep=0in]
    \item SA-Net~\cite{soans2020sa}, which is a single-view state-action pair recognition method; 
    \item A two-stream VGG-Net~\cite{Rezazadegan17:Action} which is an action recognition pipeline; and
    \item An ablation of \mvsanet{} wherein we substitute the gating network with a single classifier and eliminate all other classifiers. The input is the set of aggregated features extracted from each viewpoint, similar to the input of the gating network. Onion status prediction is determined through majority voting in the multi-view baselines. 
\end{enumerate}

\begin{table}[ht!]
\fontsize{6.4}{6}\selectfont
\addtolength{\tabcolsep}{-4pt}
\setlength\extrarowheight{3pt}
\caption{Mean classification accuracy (\%) on the onion-sorting task. Note that `-----' denotes not applicable.}
\begin{center}
\begin{tabular}{c|cccc}
\hline
{\multirow{2}{*}{\textbf{Method}}}&\textbf{Onion}&\textbf{End-effector}&\textbf{Onion}&{\multirow{2}{*}{\textbf{Action}}}\\
&\textbf{Location}&\textbf{Location}&\textbf{Status}&\\
\hline
Single-view SA-Net (front view) &85.70$\pm$0.1&91.02$\pm$0.1&40.06$\pm$0.2&93.70$\pm$0.1\\

Single-view SA-Net (side view) &86.24$\pm$0.1&90.83$\pm$0.1&39.40$\pm$0.2&91.44$\pm$0.1\\

Single-view SA-Net (top-down view) &72.09$\pm$0.3&72.76$\pm$0.3&59.27$\pm$0.7&80.07$\pm$0.2\\

Two-Stream VGG-Net (front view)&-----&-----&-----&80.92$\pm$0.1\\

Two-Stream VGG-Net (side view)&-----&-----&-----&76.26$\pm$0.2\\

Two-Stream VGG-Net (top-down view)&-----&-----&-----&69.51$\pm$0.2\\

Two-Stream VGG-Net (multi-view)&-----&-----&-----&80.92$\pm$0.1\\

\mvsanet{} without gating network & 87.15$\pm$0.1&92.11$\pm$0.1& 64.24$\pm$0.2& 95.66$\pm$0.1\\

\textbf{\mvsanet{}}&\textbf{89.84$\pm$0.1}&\textbf{95.68$\pm$0.1}&\textbf{68.21$\pm$0.2}&\textbf{97.67$\pm$0.1}\\
\hline
\end{tabular}
\label{tab:table_1}
\end{center}
\vspace{-4mm}
\end{table}

\begin{table}[ht!]
\fontsize{6.9}{6}\selectfont
\addtolength{\tabcolsep}{-4pt}
\setlength\extrarowheight{3pt}
\caption{Evaluation on the patroller-attacker experiment.}
\begin{center}
\begin{tabular}{c|cccc}
\hline
\textbf{Method}&\textbf{X}&\textbf{Y}&\textbf{$\theta$}&\textbf{Action}\\
\hline
Single-view SA-Net (Cam 1) &19.72$\pm$0.1&16.58$\pm$0.1&18.29$\pm$0.6&18.37$\pm$0.1\\
Single-view SA-Net (Cam 2)&59.87$\pm$0.3&67.04$\pm$0.3&64.60$\pm$0.5&65.93$\pm$0.2\\
Two-Stream VGG-Net (Cam 1)&-----&-----&-----&14.18$\pm$0.4\\
Two-Stream VGG-Net (Cam 2)&-----&-----&-----&52.95$\pm$0.9\\
Two-Stream VGG-Net (multi-view)&-----&-----&-----&84.86$\pm$0.4\\
\mvsanet{} without gating network &90.48$\pm$0.1&92.91$\pm$0.2&93.12$\pm$0.2&93.09$\pm$0.2\\
\textbf{\mvsanet{}}&\textbf{94.66$\pm$0.1}&\textbf{97.03$\pm$0.1}&\textbf{97.88$\pm$0.2}&\textbf{96.72$\pm$0.1}\\
\hline

\end{tabular}
\label{tab:table_2}
\end{center}
\vspace{-3mm}
\end{table}

\vspace{0.1in}
\noindent\textbf{Formative Evaluation.}~
In both domains, we assess classification accuracy through 5-fold cross-validation and compare \mvsanet{}'s performance against the baseline methods. 
Table~\ref{tab:table_1} shows that the single-view models perform poorly, particularly in detecting the onion status. \mvsanet{} outperforms all single-view pipelines by a considerable margin in both state and action recognition. Thus, the multi-view setup is indeed significantly more robust than single views. \mvsanet{} also achieves superior accuracy compared to the multi-view ablation that does not use the gating networks, demonstrating the effectiveness of incorporating the gating network into the architecture. 
We present analogous performance comparisons in the patrolling domain in Table~\ref{tab:table_2}, where Cam 2 clearly provides more informative results than Cam 1, and \mvsanet{} effectively leverages both views. 

\vspace{0.1in}
\noindent\textbf{Ablation and Robustness Study.}~
We conduct ablation experiments to demonstrate the efficacy of the depth channel and the importance of the connections between the state and action networks. We also evaluate the robustness of \mvsanet{} in scenarios where relatively high levels of noise occur due to unreliable sensing. 
For the robustness tests, we created two datasets: one containing noisy RGB images and the other containing overexposed images. These form one of the multiple streams from the camera. As shown in Tables~\ref{tab:table_3} and~\ref{tab:table_4}, \mvsanet{} successfully retrieves valuable information from the cameras that remain unaffected by noise or extreme brightness, maintaining its high performance in these realistic scenarios. 

\begin{table}[htbp]
\fontsize{6.3}{6}\selectfont
\addtolength{\tabcolsep}{-4pt}
\setlength\extrarowheight{3pt}
\caption{Performance of ablated \mvsanet{} and under noisy input and bad lighting settings for the onion-sorting task. Note that adding noise completely distorts onion status prediction.}
\begin{center}
\begin{tabular}{c|cccc}
\hline
{\multirow{2}{*}{\textbf{Method}}}&\textbf{Onion}&\textbf{End-effector}&\textbf{Onion}&{\multirow{2}{*}{\textbf{Action}}}\\
&\textbf{Location}&\textbf{Location}&\textbf{Status}&\\
\hline

\textbf{Architecture Ablations} &&&&\\
\hline

\textbf{\mvsanet{}} w/o connection&{\multirow{2}{*}{89.84$\pm$0.1}}&{\multirow{2}{*}{95.68$\pm$0.2}}&{\multirow{2}{*}{68.21$\pm$0.3}}&{\multirow{2}{*}{93.20$\pm$0.1}}\\
between state \& action networks&&&&\\

\textbf{\mvsanet{}} w/o depth channel&83.96$\pm$0.7&87.90$\pm$0.9&68.21$\pm$1.0&89.78$\pm$0.7\\

\hline
\textbf{Robustness to Noise} &&&&\\
\hline

SA-Net w/ noise (front view) &62.79$\pm$1.5&68.44$\pm$1.2&0.0$\pm$0.0&79.23$\pm$1.1\\

Multi-view ablation w/ noise &86.05$\pm$0.3&87.71$\pm$0.3&62.58$\pm$0.4&90.87$\pm$0.2\\

\textbf{\mvsanet{}} w/ noise &\textbf{88.70$\pm$0.2}&\textbf{89.69$\pm$0.3}&\textbf{65.23$\pm$0.3}&\textbf{92.97$\pm$0.2}\\

\hline
\textbf{Robustness to Bad Lighting} &&&&\\
\hline

SA-Net w/ bad lighting &{\multirow{2}{*}{84.17$\pm$0.1}}&{\multirow{2}{*}{89.20$\pm$0.2}}&{\multirow{2}{*}{31.79$\pm$1.5}}&{\multirow{2}{*}{90.51$\pm$0.2}}\\
(front view)&&&&\\

\mvsanet{} without gating network &{\multirow{2}{*}{86.36$\pm$0.1}}&{\multirow{2}{*}{90.02$\pm$0.2}}&{\multirow{2}{*}{68.02$\pm$0.2}}&{\multirow{2}{*}{92.88$\pm$0.2}}\\
w/ bad lighting&&&&\\

\textbf{\mvsanet{}} w/ bad lighting &\textbf{89.04$\pm$0.1}&\textbf{90.70$\pm$0.1}&\textbf{68.02$\pm$0.2}&\textbf{92.88$\pm$0.2}\\
\hline

\end{tabular}
\label{tab:table_3}
\end{center}
\vspace{-3mm}
\end{table}


\begin{table}[htbp]
\fontsize{6.3}{6}\selectfont
\addtolength{\tabcolsep}{-4pt}
\setlength\extrarowheight{3pt}
\caption{Ablation performance and under noisy streams for the Patrolling domain.}
\begin{center}
\begin{tabular}{c|cccc}
\hline
\textbf{Method}&\textbf{X}&\textbf{Y}&\textbf{$\theta$}&\textbf{Action}\\
\hline

\textbf{Architecture Ablations} &&&&\\
\hline

\textbf{\mvsanet{}} w/o connection &{\multirow{2}{*}{94.66$\pm$0.2}}&{\multirow{2}{*}{97.03$\pm$0.3}}&{\multirow{2}{*}{97.88$\pm$0.2}}&{\multirow{2}{*}{91.72$\pm$0.3}}\\
between state \& action networks&&&&\\

\textbf{\mvsanet{}} w/o depth channel &86.49$\pm$0.7&90.81$\pm$0.7&92.33$\pm$1.8&90.26$\pm$1.0\\

\hline
\textbf{Robustness to Noise} &&&&\\
\hline

SA-Net w/ noise (Cam 1) &14.45$\pm$1.0&12.64$\pm$0.8&14.90$\pm$2.1&15.53$\pm$1.9\\

Multi-view ablation w/ noise&88.29$\pm$0.5&89.87$\pm$0.5&92.33$\pm$0.9&90.45$\pm$1.2\\

\textbf{\mvsanet{}} w/ noise &\textbf{90.36$\pm$0.5}&\textbf{92.83$\pm$0.3}&\textbf{95.13$\pm$0.9}&\textbf{93.08$\pm$0.9}\\

\hline
\textbf{Robustness to Bad Lighting} &&&&\\
\hline

SA-Net w/ bad lighting (Cam 1) &19.04$\pm$1.2&15.78$\pm$1.1&16.98$\pm$1.9&16.94$\pm$1.7\\

\mvsanet{} without gating network &{\multirow{2}{*}{89.90$\pm$0.4}}&{\multirow{2}{*}{90.52$\pm$0.3}}&{\multirow{2}{*}{92.66$\pm$0.7}}&{\multirow{2}{*}{92.07$\pm$0.9}}\\
w/ bad lighting&&&&\\

\textbf{\mvsanet{}} w/ bad lighting &\textbf{92.85$\pm$0.4}&\textbf{95.11$\pm$0.3}&\textbf{96.29$\pm$0.4}&\textbf{94.05$\pm$0.7}\\
\hline

\end{tabular}
\label{tab:table_4}
\end{center}
\vspace{-4mm}
\end{table}


\noindent\textbf{Evaluating \mvsanet{} toward LfO.}
In order to assess the effectiveness of the state-action trajectories provided by \mvsanet{}, we use these trajectories as input for the state-of-the-art inverse RL algorithm, MAP-BIRL~\cite{choi2011map}. In our evaluation, we employ learned behavior accuracy (LBA) to compare the learned policy with the true expert policy to determine the quality of the learned policy. To achieve this, we conduct 10 independent trials of onion sorting yielding 10 trajectories each comprising 40$\pm$5 steps of state-action pairs. Two different human subjects performed the sorting task (each performed 5 trials) to enhance the model's robustness. {\em Utilizing the predictions provided by \mvsanet{} resulted in an LBA of 97.9\% as compared to 83.3\% using the single-view SA-Net on the top-down view.} Therefore, \mvsanet{}'s multi-view fusion generates state-action trajectories that result in significantly improved LfO performance compared to single-view trajectories. This improvement is attributed to its robustness in handling occlusion and inherent sensor noise.  
As such, we provide evidence that \mvsanet{} can be directly applied in real-world robotic domains, leading to substantially improved outcomes ({\bf please refer to the supplementary video for more details.})\\

\vspace{-0.40cm}
\section{Challenges and Future Work}
\label{sec:challenges_and_future_work}
During the development of our pipeline, a few practical considerations came to light, though they are often just part of the typical development landscape. The future, meanwhile, holds promise for even more enhanced capabilities.\\
\noindent\textbf{Challenges.}  
Operating multiple cameras on a single system poses a significant demand for hardware. This necessitates robust and optimized hardware to ensure smooth and uninterrupted functioning. We solved this by distributing the process into multiple machines. Our model mandates that all cameras transmit their data concurrently. However, achieving this synchronicity is challenging due to inevitable camera latencies and the varied processing speeds of heterogeneous camera systems. To handle this, we fixate the capturing speed to 10 FPS to minimize the processing latency. 

\noindent\textbf{Future Work.}
The current model recognizes actions within a set number of frames. Enhancing the model's adaptability by identifying action duration before recognition is advantageous. This approach detects actions beyond the fixed-frame limit, increasing flexibility and precision. We aim to further adapt our system for varying camera angles during training and testing, through data augmentation and diverse training data. Future improvements could involve transforming the model's structure and substituting the classification network with a regression model for continuous state-action predictions.

In sum, while our current model stands as a significant contribution to the domain, there remains much room for growth and adaptation.

\section{Conclusions}
\label{sec:conclusion}

We introduce \mvsanet{}, a novel architecture that handles inputs from multiple viewpoints to articulate desiderata while learning a robust model. Tailored meticulously for dependable robotic systems, \mvsanet{} improves upon single-view architectures by leveraging information from multiple viewpoints, enabling accurate inference with noisy input streams. While using multiple viewpoints is common in literature, our novelty lies in performing state and action recognition from various perspectives. The results show our approach surpasses baselines in prediction accuracy, potentially enabling more effective deployment in noisy environments. Future enhancements may focus on adaptive action recognition, camera flexibility, and continuous predictions.

\section*{ACKNOWLEDGMENT}

We thank Prasanth Suresh for providing the MAP-BIRL baseline for IRL evaluations and experimentation assistance. This work was enabled in part by NSF grant \#IIS-1830421 and a Phase 1 grant from the GA Research Alliance to PD.

\bibliographystyle{abbrv}
\bibliography{shdaCoRLWkshp23}

\end{document}